\documentclass[9.5pt, journal]{IEEEtai}

\usepackage[colorlinks,urlcolor=blue,linkcolor=blue,citecolor=blue]{hyperref}

\usepackage{color,array}
\ifCLASSINFOpdf
\else
\fi
\usepackage{array}
\usepackage[hang]{footmisc}

\usepackage[subrefformat=parens,labelformat=parens,caption=false,font=footnotesize]{subfig}
\usepackage{amsmath}
\usepackage[nolist,nohyperlinks]{acronym}
\usepackage{amsmath}
\DeclareMathOperator*{\argmax}{arg\,max}

	\begin{acronym}
    \acro{4G}{fourth generation}
    \acro{5G}{fifth generation}
    \acro{AoA}{angle of arrival}
   	\acro{AoD}{angle of departure}
   	\acro{AP}{access point}
    \acro{BCRLB}{Bayesian CRLB}
    \acro{BS}{base stations}
	\acro{CDF}{cumulative density function}
    \acro{CF}{closed-form}
    \acro{CRLB}{Cramer-Rao lower bound}
    \acro{ECDF}{empirical cumulative distribution function}
    \acro{EI}{Exponential Integral}
    \acro{eMBB}{enhanced mobile broadband}
    \acro{FIM}{Fisher Information Matrix}
    \acro{GoF}{goodness-of-fit}
    \acro{GPS}{global positioning system}
    \acro{GNSS}{global navigation satellite system}
    \acro{HetNets}{heterogeneous networks}
    \acro{LOS}{line of sight}
    \acro{MAB}{multi-armed bandit}
    \acro{MBS}{macro base station}
    \acro{MEC}{mobile-edge computing}
    \acro{MIMO}{multiple input multiple output}
    \acro{mm-wave}{millimeter wave}
    \acro{mMTC}{massive machine-type communications}
    \acro{MS}{mobile station}
    \acro{MVUE}{minimum-variance unbiased estimator}
    \acro{NLOS}{non line-of-sight}
    \acro{OFDM}{orthogonal frequency division multiplexing}
    \acro{PDF}{probability density function}
    \acro{PGF}{probability generating functional}
    \acro{PLCP}{Poisson line Cox process}
    \acro{PLT}{Poisson line tessellation}
    \acro{PLP}{Poisson line process}
    \acro{PPP}{Poisson point process}
    \acro{PV}{Poisson-Voronoi}
    \acro{QoS}{quality of service}
    \acro{RAT}{radio access technique}
    \acro{RL}{reinforcement-learning}
    \acro{RSSI}{received signal-strength indicator}
    \acro{BS}{base station}
   	\acro{SINR}{signal to interference plus noise ratio}
    \acro{SNR}{signal to noise ratio}
    \acro{TS}{Thompson Sampling}
    \acro{TS-CD}{TS with change-detection}
    \acro{KS}{Kolmogorov-Smirnov}
    \acro{UCB}{upper confidence bound}
	\acro{ULA}{uniform linear array}
	\acro{UE}{user equipment}
 	\acro{URLLC}{ultra-reliable low-latency communications}
    \acro{V2V}{vehicle-to-vehicle}    
\end{acronym}

\usepackage{graphicx,wrapfig,lipsum}
\usepackage{rotating}
\usepackage{amsmath, amssymb, amsthm}
\usepackage{stmaryrd}
\usepackage{tikz}
\usepackage{verbatim}
\usepackage{url}
\usepackage[utf8]{inputenc}
\usepackage[english]{babel}
\newtheorem{theorem}{Theorem}[]

\newtheorem{proposition}{Proposition}[]
\newtheorem{lemma}[]{Lemma}
\usepackage{multicol}
\usepackage{xr-hyper}

\usepackage{algorithm}
\usepackage{algpseudocode}
\usepackage{scalerel}
\usepackage{multicol}
\usepackage{setspace}
\newtheorem{definition}{Definition}
\usepackage[noadjust]{cite}
\usepackage{booktabs}

\usepackage{bbm}
\usepackage[normalem]{ulem}
\usepackage{color}
\usepackage{dsfont}
\usepackage{bm}
\usepackage{setspace}
\usetikzlibrary{automata}
\usetikzlibrary{arrows}
\usetikzlibrary{shapes.geometric, arrows}
\usepackage[]{footmisc}
\usetikzlibrary{positioning}
\usetikzlibrary{calc}
\usetikzlibrary{arrows}
\allowdisplaybreaks

\newcommand\blfootnote[1]{%
  \begingroup
  \renewcommand\thefootnote{}\footnote{#1}%
  \addtocounter{footnote}{-1}%
  \endgroup
}
\usepackage{mathtools}

\usepackage{xcolor}
\usepackage{scalerel}
\usepackage{float}
\usepackage{tikz}
\usetikzlibrary{svg.path}

\usepackage{tikz,xcolor,hyperref}

\definecolor{lime}{HTML}{A6CE39}
\DeclareRobustCommand{\orcidicon}{%
	\begin{tikzpicture}
	\draw[lime, fill=lime] (0,0) 
	circle [radius=0.16] 
	node[white] {{\fontfamily{qag}\selectfont \tiny ID}};
	\draw[white, fill=white] (-0.0625,0.095) 
	circle [radius=0.007];
	\end{tikzpicture}
	\hspace{-2mm}
}

\foreach \x in {A, ..., Z}{%
	\expandafter\xdef\csname orcid\x\endcsname{\noexpand\href{https://orcid.org/\csname orcidauthor\x\endcsname}{\noexpand\orcidicon}}
}

\begin{document}

	\title{Kolmogorov–Smirnov Test-Based Actively-Adaptive Thompson Sampling for Non-Stationary Bandits}
	\author{Gourab Ghatak\orcidA{},~\IEEEmembership{Member,~IEEE,}  \thanks{G. Ghatak is with the Department of ECE, IIIT Delhi, India. Email: \texttt{gourab.ghatak@iiitd.ac.in}} Hardhik Mohanty\orcidB{}\thanks{H. Mohanty is with the Department of Electrical Engineering, IIT Kharagpur, India. Email:~\texttt{hardhikiitkgp2018@iitkgp.ac.in} }, and Aniq Ur Rahman\orcidC{} \thanks{A. U. Rahman is with the CEMSE Division, King Abdullah University of Science and Technology (KAUST), Thuwal 23955, Saudi Arabia. He contributed to this work before joining KAUST as a student. Email:~\texttt{aniqur.rahman@kaust.edu.sa} } \vspace{-0.8cm}}

\setcounter{page}{1}
	\maketitle
		
\blfootnote{The source code for this work is available for download \cite{github_gof}.}

\begin{abstract}
We consider the non-stationary multi-armed bandit (MAB) framework and propose a Kolmogorov-Smirnov (KS) test based Thompson Sampling (TS) algorithm named \texttt{TS-KS}, that actively detects change points and resets the TS parameters once a change is detected. In particular, for the two-armed bandit case, we derive bounds on the number of samples of the reward distribution to detect the change once it occurs. Consequently, we show that the proposed algorithm has sub-linear regret. Contrary to existing works, our algorithm is able to detect a change when the underlying reward distribution changes even though the mean reward remains the same. {Finally, to test the efficacy of the proposed algorithm, we employ it in two case-studies: i) task-offloading scenario in wireless edge-computing, and ii) portfolio optimization.} Our results show that the proposed \texttt{TS-KS} algorithm outperforms not only the static TS algorithm but also it performs better than other bandit algorithms designed for non-stationary environments. {Moreover, the performance of \texttt{TS-KS} is at par with the state-of-the-art forecasting algorithms such as \texttt{Facebook-PROPHET} and \texttt{ARIMA}.}
\end{abstract}

\begin{IEEEImpStatement}
The applications of the multi-armed bandit (MAB) framework is well-established in literature. However, the non-stationary MAB problem is notoriously difficult to analyze mathematically. Contrary to the existing works, the proposed algorithm in this paper can detect a change in the environment when the underlying reward distribution changes even though the mean remains constant. {In the task-offloading application in wireless edge-computing, and in the portfolio optimization problem, the proposed algorithm outperforms the contending algorithms across all time-windows.} This work will find significant applications in epidemiology, e.g., in randomized clinical trials, wireless networks, computational advertisement, and financial portfolio design.
\end{IEEEImpStatement}

\begin{IEEEkeywords}
Multi-armed bandits, Thompson sampling, Kolmogorov-Smirnov test, edge-computing, 5G.
\end{IEEEkeywords}

\section{Introduction}
\IEEEPARstart{I}n \ac{RL}, the \ac{MAB} is a framework to facilitate sequential decision-making, wherein, a player selects a subset possible actions based on the rewards of previous action selections~\cite{lee2020optimization, bouneffouf2019survey, englert2016combined, burtini2015survey}. The \ac{MAB} framework has found applications in the fields of load demand sensing~\cite{lesage2017multi}, online recommendation systems~\cite{li2016collaborative}, computational advertisement~\cite{buccapatnam2017reward}, and wireless communications~\cite{rahman2019beam}, among others. {The \ac{MAB} algorithms are particularly useful since they can directly be employed in an uncertain environment without any training phase.} In the basic \ac{MAB} setting, an agent (or player) chooses one out of $K$ independent choices (or arms), and obtains an associated reward. Each arm corresponds to a (possibly non-identical) reward distribution, that is unknown to the player. The objective of the player is to device an algorithm which minimizes the difference between the total rewards of the player and the highest expected reward, over a period of time, called the horizon $T$. To derive such an algorithm, the player maintains a belief associated to each choice as the temporal dynamics of the \ac{MAB} setting evolves. Consequently, at each time-slot, the player selects the choice with the highest belief. This action-selection procedure presents an exploration-exploitation dilemma. In the classical setting, the reward distributions remain stationary over the entire time of play. However, most practical systems are characterized by a non-stationary reward framework, which is notoriously difficult to analyze analytically.
{Additionally, classical bandit algorithms are designed to select the arm with the highest mean reward. On the contrary, for several use-cases, such as portfolio optimization and clinical trials, an optimal arm may not be the one with the highest mean reward, but the one that achieves the best risk-reward balance. Recently, Zhu {\it et al.}~\cite{zhu2020thompson} have developed modified TS algorithms for mean-variance bandits and provide comprehensive regret analyses for Gaussian and Bernoulli bandits. Indeed, for such use-cases, an algorithm that yields a lower expected payout but is less {\it risky} may be preferable. Since risk is characterized by the variance of the reward, a detection of the change in the \ac{CDF} of the reward distribution becomes imperative.} In this work, we propose and characterize a \ac{KS} test based bandit algorithm that provides asymptotic regret guarantees with an arbitrary probability.

{\bf Related Work:}
For stationary settings, several bandit algorithms, e.g., see~\cite{contal2013parallel} perform optimally. On the contrary, the \ac{TS} algorithm~\cite{thompson1933likelihood}, has gotten considerable interest recently due to its remarkably augmented performance vis-a-vis index-based policies~\cite{granmo2010solving}. However, the analytical characterization of the \ac{TS} algorithm was challenging due to its randomized nature, until the work by Agarwal and Goyal~\cite{agrawal2012analysis}.

In case of non-stationary reward distributions, most bandit algorithms loose mathematical tractability. For such non-stationary environments, researchers generally employ either of the two approaches: i) passively adaptive policies (e.g., see~\cite{garivier2008upper, besbes2014stochastic}) which remain oblivious to the change instants and gradually decrease the weights given to past rewards, and ii) actively adaptive policies (e.g., see~\cite{hartland2006multi, yu2009piecewise, cao2019nearly}) which track the changes and reset the algorithm on the detection of a change. Generally speaking, for the passively adaptive algorithms, rigorous performance bounds can be derived. {Among the passively adaptive policies, Garivier and Moulines~\cite{garivier2008upper} have considered a scenario where the distribution of the rewards remain constant over epochs and change at unknown time instants. They have analyzed the theoretical upper and lower bounds of regret for the discounted \ac{UCB} (D-UCB) and sliding window \ac{UCB} (SW-UCB). Besbes {\it et al.}~\cite{besbes2014stochastic} have developed a near optimal policy, REXP3. The authors have established lower bounds on the performance of any non-anticipating policy for a general class of non-stationary reward distributions. Following this, the optimal exploration exploitation trade-off for non-stationary environments were studied by the authors in~\cite{besbes2019optimal}.} 

On the other hand, actively adaptive algorithms outperform passively adaptive ones, as has been shown experimentally in e.g.,~\cite{mellor2013thompson}. Pertaining to the actively adaptive policies, Hartland et al.~\cite{hartland2006multi} have proposed an algorithm \texttt{Adapt-EVE} for {abrupt changes} in the environment. The change is detected via the Page-Hinkley statistical test (PHT). However, their evaluation is empirical and deviod of any performance bounds. PHT has found other applications, e.g., to adapt the window length of SW-UCL~\cite{srivastava2014surveillance}. Furthermore, the works by Yu {\it et. al}~\cite{yu2009piecewise} and Cao {\it et al.}~\cite{cao2019nearly} detect a change in the empirical means of the rewards of the arms. However, in both these works, the authors assume a fixed number of changes withing an interval to derive the respective bounds. The work that is closest to ours is~\cite{9194367}, where the author assumed a change-detection based \ac{TS} algorithm. However, similar to \cite{yu2009piecewise, cao2019nearly}, the author has proposed a change detection mechanism based on the comparison of means. On the same line, recently, Auer {\it et al.}~\cite{auer2019adaptively} have proposed the algorithm \texttt{ADSWITCH} which is also based primarily on mean-estimation based checks for all the arms, and achieves nearly-optimal regret bounds without the prior knowledge of the total number of changes. {Remarkably, the authors show regret guarantees for \texttt{ADSWITCH} without any pre-condition on the quanta of changes with each change point.} However, such an algorithm fails to detect the change when the underlying distribution changes while the mean remains the same. {Thus, there is a need to develop a MAB framework that is sensitive to change in the higher moments of the reward distribution. Such a setting can then be applied to the context of randomized clinical trial~\cite{villar2015multi} in which the characteristic of the targeted pathogen changes abruptly (e.g., due to mutations), or, e.g., in computational advertisement in order to detect and adapt to a change in user profile~\cite{jahanbakhsh2020applying}.} On the contrary, in this work, we propose a detection mechanism based on the change of the \acp{CDF}. In particular:
\begin{itemize}
    \item We introduce a novel change detection framework for the non-stationary \ac{MAB} problem based on the \ac{KS} goodness-of-fit test. The proposed algorithm detects a change when the underlying reward distribution changes even when the mean remains the same {(e.g., \cite{auer2019adaptively,cao2019nearly,yu2009piecewise})}.
    \item Based on the \ac{KS} test based change detection, we propose a actively adaptive \ac{TS} algorithm called \texttt{TS-KS}, which resets the parameters of the classical \ac{TS} algorithm on detection of the change. Furthermore, by considering a change in the mean of the Gaussian distributed rewards, we derive lower bounds on i) the number of samples of the reward distribution of an arm to accurately estimate its \ac{ECDF}, and ii) the number of samples $n_T$ from the reward distribution. Consequently, we derive a bound on the regret of the \texttt{TS-KS} algorithm ans show that it has a sub-linear regret with an arbitrary probability. We also investigate the conditions under which the number of samples required by \texttt{TS-KS} is smaller than the state-of-the-art mean-estimation based change-detection methods.
    \item Finally, based on the \texttt{TS-KS} algorithm we present {two case-studies. First, we investigate a task-offloading scenario} in edge-computing where a secondary user in the network selects one out of a set of servers to offload a computational task in the presence of primary users. {Then, we employ the proposed \texttt{TS-KS} algorithm in the portfolio optimization problem.} Our result highlights that the proposed algorithm outperforms not only the classical \texttt{TS} algorithm, but also it perform better than other passively-adaptive and actively-adaptive TS algorithms that designed for non-stationary environments.
\end{itemize}

\section{The Two-Armed Non-Stationary Bandit Setting}
In this paper we study the two-armed bandit framework\footnote{Although the framework developed in this paper can be extended to ${K}$ arms, for tractability, we restrict our analysis to ${K} = 2$.}, with arms $a_i$, where $i \in \{1,2\}$. This framework is the same as the one considered in \cite{9194367}. The player selects an arm at each time step $n$, and obtains a reward $R_{a_i}(n)$. The reward of arm $a_i$ is assumed to follow a Gaussian distribution\footnote{Gaussian distribution has been previously considered in bandit literature, e.g.,~\cite{srivastava2014surveillance}.}, i.e., $R_{a_i}(n) \sim \mathcal{N}\left(\mu_i(n),\sigma^2\right)$, with non-stationary mean that is unknown to the player. Moreover, for the purpose of the mathematical analysis, we assume a fixed variance $\sigma^2$ across arms, although the framework is applicable in practice for the case when the variances are distinct across arms and piece-wise non-stationary. Additionally, let us consider the following lower bound on the difference between the mean reward of the two arms:
\begin{align}
  \left| \mu_i(n) - \mu_j(n) \right| \geq \Delta_\mu, \quad {i \neq j}, \forall n. \label{eq:mindelmu}
\end{align}

%Without loss of generality, we assume that $\mu_{l}(n) \geq \mu_{l+1}(n)$ at $n = 0$, i.e., $\mu_{l}(0) \geq \mu_{l+1}(0) + \Delta_\mu$. It must be noted that the analysis remains the same in the case when $\mu_{l}(n) < \mu_{l+1}(n)$. 
In our work, we consider a piecewise-sationary environment in which {the values of $\mu_i(n)$ change at random time-instants, $T_{C_l}$, $l = 1, 2, \ldots$, with $T_{C_0}$ is at $n = 0$.} In particular, we assume that the change instants follow a Poisson arrival process with parameter $\lambda_C$. Thus, the quantity {$T_{C_{l+1}} - T_{C_{l}}$ is exponentially distributed: 
\begin{align}
        \mathbb{P}\left(T_{C_{l+1}} - T_{C_{l}} \leq k\right) = 1 - \exp\left(-\lambda_C k\right).
        \label{eq:change} 
\end{align}
{Thanks to the memoryless property of the exponential random variable, each arrival is stochastically independent to all the other arrivals in the process. {Thus, in contrast to the existing studies that consider a fixed number of changes (e.g., see~\cite{yu2009piecewise, cao2019nearly, auer2019adaptively}), we consider a framework with a random number of changes governed by the intensity parameter $\lambda_C$. This reflects in the parameter $\lambda_C$ appearing in the regret bound of our algorithm as compared to e.g., the exact number of changes in \cite{auer2019adaptively}. The case with a fixed number of changes is thus a realization of the Poisson arrival process that occur with a Poisson probability distribution function.}}
Furthermore, we assume that both the arms change simultaneously at $T_{C_l}$. {Thus, in contrast to the existing studies that consider a fixed number of changes (e.g., see~\cite{yu2009piecewise, cao2019nearly, auer2019adaptively}), we study a framework consisting of a random number of changes governed by the intensity parameter $\lambda_C$. Owing to the independent scattering property of the Poisson arrival process, the presented framework models a {\it completely random} change-point arrival process.}
}
  Furthermore, we assume that the mean rewards of the same arm across different time slots is bounded as:
\begin{align}
  \left|\mu_i(n) - \mu_i(k) \right| \in \{0\} \cup \left[\Delta_{\min},\Delta_{\max}\right], \quad n \neq k. 
  \label{eq:mindel}
\end{align}
In particular, either the mean of the reward of an arm does not change, or it changes uniformly by a magnitude bounded between $\Delta_{\min}$ and $\Delta_{\max}$. {Note that as discussed later in Proposition 1, the KS-test based change detection works even when $\Delta_{\min} = 0$ as long as $\sigma_2 \neq \sigma_1$. However, we consider a change in the reward distribution to be characterized by a change in the corresponding means so as to enable a fair comparison with the state-of-the-art mean-estimation based actively adaptive bandit algorithms.}

{At this point, it can be noted that the minimum difference in the mean reward before and after a change point, i.e., $\Delta_{\min}$ is equivalent to $2 \epsilon$ in \cite{yu2009piecewise} and $\delta_k$ in \cite{cao2019nearly}.}
% \vspace{-0.3cm}

{\bf Regret}
The function of an algorithm $\pi$ is to make a decision at each time step $n$, regarding $a_{\pi}(n)$, i.e., which arm to play. Consequently, $\pi$ receives a reward $R_{a_\pi}(n) \sim \mathcal{N}\left(\mu_{a_\pi}(n), \sigma^2\right)$. To characterize the regret performance of the algorithm, let at a time-step $n$, the optimal arm be denoted by $a_i$. Then, the regret after $T$ rounds of action-selection is given by:
\begin{align}
    \mathcal{R}(T) = \sum_{n=0}^T \left(\mu^*(n) - \mu_{a_{\pi}}(n)\right). \nonumber 
\end{align}
The aim is to devise a policy $\pi$ that achieves a bound on the expected regret: $\mathbb{E} \left[\mathcal{R}(T)\right]$. In this regard, in the next section we discuss the proposed KS-test based change-detection mechanism that tracks changes in the environment.

\begin{algorithm}
\caption{\texttt{TS-KS}}
\begin{algorithmic}[1]
    \State \textbf{Parameters:} $n_T$, $N$, $t_{ref}$, $T$.  
    \State \textbf{Initialization:} Count = 1, $\mathcal{S}_j = \{\}$ $\forall{j} \in \{1, 2\}$ $\alpha_j = 1, \beta_j = 1$ $\forall{j} \in \{1, 2\}$
    \For {$t = 1, 2, \ldots, T$}
    \State $\theta_j \sim Beta(\alpha_j, \beta_j) \forall j \in \{1, 2\}$  
    \State   {Play arm $a_k$ such that $k = \operatorname{argmax_j} \theta_j$}
    \State Observe reward $\Tilde{r_t}$ 
    \State $\mathcal{S}_k = \mathcal{S}_k \cup \Tilde{\{r_t\}}$ \Comment{Store reward}
    \State $R_\pi(t) \gets  \dfrac{\Tilde{r_t} - \mu_{\min} - \sigma Q^{-1}\left(\epsilon_b\right)}{2\sigma Q^{-1}\left(\epsilon_b\right) + \mu_{\max} - \mu_{\min}}$
    \State Obtain $x_t \sim {\rm Bernoulli}(\Tilde{R_\pi(t)})$ \Comment{Bernoulli Trial}
    \State Update $\alpha_k = \alpha_k + x_t$ and $\beta_k = \beta_k + 1 - x_t$ 
    \State Count = Count + 1
    \If {{Count $> (T_N + n_T)$}}
        \State Identify optimal arm $k_b \doteq \operatorname{argmax_j} \frac{\alpha_j}{\alpha_j + \beta_j}$
        \State $\mathcal{S}_t = \cup_{i = L-n}^L \mathcal{S}_{k_b}(i)$ \Comment{Test Sequence}
        \State $\mathcal{S}_e = \cup _{i = L-n-N}^{L-n} \mathcal{S}_{k_b}(i)$ \Comment{Estimate Sequence}
        \State{$t_{stat}$ = \texttt{KS2Samp}($S_e$, $S_t$)} \Comment{K-S Test}
        \If {$t_{stat} > t_{ref}$}
            \State Count = 1 \Comment{Change Detected}
            \State $\alpha_j = 1, \beta_j = 1$ $\forall{j} \in \{1, 2\}$ \Comment{Reset}
        \EndIf
    \EndIf
    \EndFor
\end{algorithmic}
\label{algo:main}
\end{algorithm}

\begin{algorithm}
\caption{\texttt{KS2Samp}}
\begin{algorithmic}[1]
    \State \textbf{Parameters:} $n$, $N$, $\mathcal{S}_t$, $\mathcal{S}_e$.
    \State $F_{\mathcal{S}_t} \gets ECDF(\mathcal{S}_t)$  \Comment{Emperical CDF of test set}
    \State $F_{\mathcal{S}_e} \gets ECDF(\mathcal{S}_e)$  \Comment{Emperical CDF of estimated set}
    \State $t_{stat} = \sqrt{\frac{nN}{n+N}}\operatorname{sup}_z |F_{\mathcal{S}_t}(z) - F_{\mathcal{S}_e}(z)|$
    \State \textbf{Return} $t_{stat}$
\end{algorithmic}
\label{algo:main}
\end{algorithm}

\section{\ac{KS} Test based \ac{TS} Framework}

In this section, we describe the proposed \ac{TS} algorithm which actively detects a change-point based on the \ac{KS} test. First let us recall the classical \ac{TS} algorithm. It assumes a Beta distribution, $\mathcal{B}(\alpha_k, \beta_k)$ to be the prior belief for the distribution of the rewards of each arm. {
    The parameters $\alpha_k, \beta_k$ are initialized to 1.} Then, at each $n$, the player generates an instance $\theta_k \sim \mathcal{B}(\alpha_k, \beta_k), \forall k \in \{1,2\}$. Consequently, it plays the choice $k = \argmax_j \theta_j$ and obtains a reward $R(n)$.
{Due to the assumption of normal distributed rewards, the values that the reward can take is in the range $R(n) \in (-\infty, \infty)$. However, for the sake of tractability and to employ the \ac{TS} algorithm of~\cite{agrawal2012analysis}, we map $R(n)$ to the range $[0,1]$~\cite{9194367}. For each arm, let us define $L$ and $U$ as: $
    \mathcal{P}(L \leq R_{a_j(n)} \leq U) \geq 1 - 2\epsilon_b, \quad \forall j \in \{1,2\}, $
where, $\epsilon_b$ is an arbitrary small positive number. Naturally, the values of $U$ and $L$ depend on the maximum and minimum values of $\mu_j$, respectively. Accordingly, let us define:
\begin{align}
    \mu_{\max} &= \max {\mu_j(n)}, \quad \forall j, n \quad \mbox{and,} \nonumber \\
    \mu_{\min} &= \min {\mu_j(n)}, \quad \forall j, n \nonumber 
\end{align}
as the bounds of the mean rewards. Consequently, $U$ and $L$ can be calculated as:
$
    U = \sigma Q^{-1}\left(\epsilon_b\right) + \mu_{\max}$ and $
    L = \mu_{\min} - \sigma Q^{-1}\left(\epsilon_b\right),$ {where $Q^{-1}(\cdot)$ is the inverse of the Gaussian $Q$-function}.
Accordingly, we define $
    R_{\pi}(n) = \frac{R_{a_j}(n) - L}{U - L}$,
which ensures that $0 \leq R_{\pi}(n) \leq 1$ with a probability $1 - 2 \epsilon_b$.} and perform a Bernoulli trial with a success probability $R_{\pi}$~\cite{agrawal2012analysis} with an outcome $x(n)$. {Of course, for the reward distributions that are naturally restricted within a finite domain (e.g., Bernoulli rewards), this step is not necessary and the rest of the algorithm follows without this step.}

The posterior distribution is consequently updated as $\alpha_k = \alpha_k + x(n)$ and $\beta_k = \beta_k + 1 - x(n)$. The sequence of the obtained rewards by the agent for each arm $j$ is used to generate a reward set $\mathcal{S}_j$. In our work, the distribution of the rewards are stationary for $T_F \geq T_N + n_T$ after each change. After $T_N$ plays of the bandit algorithm, the change-detection mechanism initiates. Each $\mathcal{S}_j$ is updated after every play for $a_j$ {(i.e., the optimal arm), until the change is detected}. In case the environment remains stationary, the beliefs of both the arms provide an accurate estimate of their respective CDFs. On the contrary, when the reward distributions of the both the arms change over time, the player has to adapt when such a change is detected. This is described in the what follows.

{\bf Active Change-point Detection: }
To actively detect the change-points, the proposed algorithm uses the two sample \ac{KS} test~\cite{berger2014kolmogorov}. In particular, the algorithm stores the rewards of the arm $a_k$ in a sequence $\mathcal{S}_k$ which is updated each time the arm $a_k$ is played. From the sequence $\mathcal{S}_k (\cdot)$, we create two sub-sequences: i) the test sequence $\mathcal{S}_t(\cdot)$: the rewards for the last $n_T$ plays of arm $j$, and ii) the estimate sequence $\mathcal{S}_e(\cdot)$: the rewards for the last $N$ rewards before the previous $n_T$ plays. In other words, if the cardinality of $\mathcal{S}_j$ is $L$, we have $\mathcal{S}_t = \cup_{i = L-n_T}^L \mathcal{S}_{k_b}(i)$ and $\mathcal{S}_e = \cup _{i = L-n_T-N}^{L-n_T} \mathcal{S}_{k_b}(i)$. Consequently, \texttt{TS-KS} constructs the \ac{ECDF} of $\mathcal{S}_t$ and $\mathcal{S}_e$, denoted by $F_{\mathcal{S}_t}(\cdot)$ and $F_{\mathcal{S}_e}(\cdot)$, respectively.

Next, we calculate the Kolmogorov distance between the two \acp{ECDF} as $D_{n_T,N} = \operatorname{sup}_z |F_{\mathcal{S}_t}(z) - F_{\mathcal{S}_e}(z)|$ and compare it with a reference test statistic value $t_{ref}$. In case $D_{n_T,N} > t_{ref}$, the test concludes that the two set of samples belong to two different probability distributions, and a change is detected. On the contrary, if $D_{n,m} < t_{ref}$, the two set of samples are considered to be i.i.d. from the same distribution.

The parameter $N$ is the minimum number of samples needed to form an accurate \ac{ECDF} of the reward distribution before the change point. On the other hand, the parameter $n_T$ is the minimum number of samples after the change-point needed to correctly detect a change. In the next section, we derive the theoretical bounds on the parameters $n_T$ and $N$. 

\section{Theoretical Bounds}
Let the reward distribution of the optimal arm before the change-point $T_{C_l}$ be denoted by $\rho_l \sim \mathcal{N}(\mu_l, \sigma^2)$ and after the change-point be denoted by $\rho_{l+1} \sim \mathcal{N}(\mu_{l+1}, \sigma^2)$. Without loss of generality, it is assumed that $\mu_{l+1} > \mu_{l}$. 
%Additionally, we consider that the difference in mean $\Delta$ of the reward distribution before and after the change-point is bounded i.e.,
%\begin{equation} 
%    \Delta_{\min} \leq |\mu_{i+1} - \mu_{i}| \leq \Delta_{\max}
%\end{equation}
%\textcolor{red}{How do we observe it to be uniformly distributed? If this is an assumption, is it necessary to have that?}
Let us denote the \ac{CDF} of $\rho_l$ by $F_l(x)$ and CDF of $\rho_{l+1}$ by $F_{l+1}(x)$. Let us mathematically characterize the {\it accuracy} of the ECDF $F_{\mathcal{S}_e}$ from $N$ observed samples. Note that contrary to \cite{9194367}, for the \texttt{TS-KS} algorithm, merely an estimate of the mean of the arm is not sufficient, and the framework must estimate the \ac{CDF} of the reward distribution. For this, let us first introduce the following definition:
{
\begin{definition}
An \ac{ECDF} $\hat{F}_X(x)$ of a random variable $X$ formed from $K$ samples is defined to be {\it accurate} with respect to its actual \ac{CDF} $F_X(x)$ if:
%\begin{equation}
%    D_K = \sqrt{K} \operatorname{sup}_x |\hat{F}_X(x) - F_X(x)| \leq t_{ref}. \nonumber 
%\end{equation}
%\end{definition}
%The above definition follows from the one-sample \ac{KS} test of $\hat{F}_X(x)$ with respect to $F_X(x)$. Thus, for the \ac{ECDF} of $\mathcal{S}_e$ to be {\it accurate}, the following condition must be satisfied:
%\begin{equation}
%    D_N = \sqrt{N} \operatorname{sup}_x |F_{\mathcal{S}_t} - F_l(x)| \leq t_{ref}.
%\end{equation}
\begin{equation}
    D_K = \operatorname{sup}_x |\hat{F}_X(x) - F_X(x)| \leq t_{ref}. \nonumber 
\end{equation}
\end{definition}
The above definition follows from the one-sample \ac{KS} test of $\hat{F}_X(x)$ with respect to $F_X(x)$. Thus, for the \ac{ECDF} of $\mathcal{S}_e$ to be {\it accurate}, the following condition must be satisfied:
\begin{equation}
    D_N = \operatorname{sup}_x |F_{\mathcal{S}_t} - F_l(x)| \leq t_{ref}.
\end{equation}}
{
\begin{proposition}
The KS-test detects a change in the bandit framework when the reward distributions change even when the mean of the rewards remain constant.
\end{proposition}
\begin{IEEEproof}
The KS statistic, defined by the supremum of the absolute value of the difference of the CDFs. Let $F_{S_t} \sim \mathcal{N}(\mu_{l},\sigma_{l}^{2})$ and $F_l \sim \mathcal{N}(\mu_{l+1},\sigma_{l+1}^{2})$, for a generic values of $\mu_l, \mu_{l+1}, \sigma_l$ and $\sigma_{l+1}$. Then, the supremum of $|F_{S_t}(x) - F_l(x)|$ is 
\begin{align}
D = &\frac{1}{2}\left(\operatorname{erf}\left(\frac{\sigma_l^2(\mu_{l+1}-\mu_l)-y_1}{\sqrt{2}\sigma_l(\sigma_l^2-\sigma_{l+1}^2)}\right)- \right. \nonumber \\
&\left.\operatorname{erf}\left(\frac{\sigma_{l+1}^2(\mu_{l+1}-\mu_l)-y_1}{\sqrt{2}\sigma_{l+1}(\sigma_l^2-\sigma_{l+1}^2)}\right)\right)
\end{align}
where, 
\begin{align}
y_{1}=&\sigma_{l} \sigma_{l+1} \left(2 \sigma_{l}^{2} \log \left(\frac{\sigma_{l}}{\sigma_{l+1}}\right)-2 \mu_{l} \mu_{l+1}- \nonumber \right. \\
&\left.2 \sigma_{l+1}^{2} \log \left(\frac{\sigma_{l}}{\sigma_{l+1}}\right)+\mu_{l}^{2}+\mu_{l+1}^{2}\right)^{\frac{1}{2}}.
\end{align}
Accordingly, for $\sigma_1 \neq \sigma_2$, the value of $D$ is non-zero even if $\mu_{l} = \mu_{l+1}$. Accordingly, for fixed mean, and a suitable threshold, the KS-test is able to detect a change in the environment which the mean-estimation based detection misses.
\end{IEEEproof}}

However, in what follows, we focus on the case when $\sigma_1 = \sigma_2 = \sigma$ and $\mu_{l} \neq \mu_{l+1}$. In the following Lemma, we bound the number of elements $N$ of $\mathcal{S}_e$ for the \ac{ECDF} of the optimal arm to be accurate with respect to its actual \ac{CDF} with a probability of at least $1 - p_{loc}$.

{
\begin{lemma}
The number of samples $N$ for which the ECDF of the optimal arm is accurate with respect to its $F_l(x)$ with probability greater than $1 - p_{loc}$ is:
\begin{align}
    N \geq \frac{1}{2t_{ref}^2}\ln \left(\frac{2}{p_{loc}}\right) \nonumber
\end{align}
\end{lemma}
\begin{IEEEproof}
We recall the Dvoretzky–Kiefer–Wolfowitz-Massart (DKWM) inequality~\cite{massart1990tight} to bound the probability that the Kolmogorov distance of the ECDF with $N$ samples to its actual CDF is greater than $t_{ref}$ as:
\begin{align}
    \mathbb{P}\left(\sup_{x \in \mathbb{R}}|{F}_{\mathcal{S}_t}(x) - F_l(x)| > t_{ref}\right) \leq 2\exp\left(-2 N t_{ref}^2\right), \nonumber \\
    \qquad \forall t_{ref} > 0.
\end{align}
Accordingly, we bound the right hand side as $2\exp\left(- 2 N t^2_{ref}\right) \leq p_{loc}$ to complete the proof.
\end{IEEEproof}
}
Next, we evaluate the number of plays of the stationary bandit framework so that at least $N$ plays of the optimal arm occurs. {First, let us note that in order to have at least one of the arms played at least $N$ times, the total number of plays is lower bounded by $T_{N} = 2N$. On the other hand, for several applications where the variance of performance of the classical \texttt{TS} algorithm remains constant, we have the following result.}
\begin{lemma}
{For the special case when the variance of the \texttt{TS} algorithm is low so that the approximation $N \approx \mathbb{E}[N]$ holds true, the} number of plays so that the optimal arm is played at least $N$ times is given by:  
{
\begin{equation*}
   T \geq T_{N} = \frac{160}{\Delta^2_\mu} \log \left(\frac{80}{\Delta^2_\mu}\right) + 2\left[\frac{48}{\Delta^4_\mu} + 18 + N\right]. \nonumber
\end{equation*}
}
\end{lemma}
\begin{IEEEproof}
For this result, first let us consider the number of plays of the \ac{TS} algorithm in the stationary regime, such that the optimal arm is played $N$ times. This in turn characterizes an {\it accurate} estimate of its \ac{CDF}. 

In the classical setting, if the \texttt{TS} algorithm is employed for $T$ plays, the mean plays of the sub-optimal $a_j$ is bounded as~\cite{agrawal2012analysis}:
$
    \mathbb{E}\left[N_j\right] \leq \frac{40 \ln T}{\Delta_\mu^2} + \frac{48}{\Delta_\mu^4} + 18. 
$
Consequently, we want the number of plays of the optimal arm to satisfy the following criterion:
{
\begin{align}
    &T - \left(\frac{40 \ln T}{\Delta_\mu^2} + \frac{48}{\Delta_\mu^4} + 18\right) \geq N, \nonumber \\
    \implies &T \geq \frac{40 \ln T}{\Delta^2_\mu} +\left[\frac{48}{\Delta^4_\mu} + 18 + N\right]
\end{align}
Then the statement of the lemma follows from applying Lemma A2 of \cite{shalev2014understanding}.
}
\end{IEEEproof}

Given the \ac{ECDF} of the reward $\rho_l$ is {\it accurate}, we characterize the performance of the change-point detection. In particular, let the parameter $n_T$ denote the minimum number of plays required to detect a change. We recall that a change is detected with $n_T$ plays when the following condition holds true:
\begin{lemma}
\label{lem:lem3}
The maximum difference between the two \acp{CDF} of $\rho_l$ and $\rho_{l+1}$ is given by $\operatorname{sup}_x |F_l(x) - F_{l+1}(x)| = \operatorname{erf}(\frac{\mu_l - \mu_{l+1}}{2\sqrt{2}\sigma})$ and the point of maximum is $x_{\max} = \frac{\mu_l + \mu_{l+1}}{2}$
\end{lemma}
{\begin{IEEEproof}
We assume the CDF before change-point follows the distribution  $F_l \sim \mathcal{N}(\mu_{l},\sigma^{2})$ and CDF after the change-point follows the distribution $F_{l+1} \sim \mathcal{N}(\mu_{l+1},\sigma^{2})$. Therefore, the difference between the CDFs is given by the following equation:
\begin{align}
    |F_l(x) - F_{l+1}(x)| =& \frac{1}{2}\left[1+\operatorname{erf}\left(\frac{x-\mu_l}{\sigma \sqrt{2}}\right)\right] - \nonumber \\
    &\frac{1}{2}\left[1+\operatorname{erf}\left(\frac{x-\mu_{l+1}}{\sigma \sqrt{2}}\right)\right] \nonumber \\
    =& \frac{1}{2}\left[\operatorname{erf}\left(\frac{x-\mu_l}{\sigma \sqrt{2}}\right) - \operatorname{erf}\left(\frac{x-\mu_{l+1}}{\sigma \sqrt{2}}\right) \right] 
\end{align}
Next to find $D_n = \operatorname{sup}_x |F_l(x) - F_{l+1}(x)|$, we first differentiate the above function with respect to $x$ in order to obtain the maximum point:
\begin{equation*}
    \frac{\partial D_n}{\partial x} = \frac{1}{\sigma \sqrt{2 \pi}} \left[e^{-\frac{1}{2}\left(\frac{x-\mu_l}{\sigma}\right)^{2}} - e^{-\frac{1}{2}\left(\frac{x-\mu_{l+1}}{\sigma}\right)^{2}} \right] = 0
\end{equation*}
%\begin{equation*}
%    \left[e^{-\frac{1}{\sqrt{2}}\left(\frac{x-\mu_l}{\sigma}\right)} + e^{-\frac{1}{\sqrt{2}}\left(\frac{x-\mu_{l+1}}{\sigma}\right)} \right] * \left[e^{-\frac{1}{\sqrt{2}}\left(\frac{x-\mu_l}{\sigma}\right)} - e^{-\frac{1}{\sqrt{2}}\left(\frac{x-\mu_{l+1}}{\sigma}\right)} \right] = 0
%\end{equation*}
%In the above equation, the first term is always positive. Hence, equating the second term to zero, we obtain:
%\begin{equation*}
%    e^{-\frac{1}{\sqrt{2}}\left(\frac{x-\mu_l}{\sigma}\right)} - e^{-\frac{1}{\sqrt{2}}\left(\frac{x-\mu_{l+1}}{\sigma}\right)} = 0
%\end{equation*}
\begin{equation*}
    \implies e^{-\frac{1}{2}\left(\frac{x-\mu_l}{\sigma}\right)^{2}} - e^{-\frac{1}{2}\left(\frac{x-\mu_{l+1}}{\sigma}\right)^{2}} = 0
\end{equation*}
Solving for $x$, we get:
% Hence, the above condition results in the following equation:
% \begin{equation*}
%     \left(\frac{x-\mu_l}{\sigma}\right)^{2} - \left(\frac{x-\mu_{l+1}}{\sigma}\right)^{2} = 0
% \end{equation*}
% Under the assumption $\mu_l \neq \mu_{l+1}$, we obtain the following equation:
% \begin{equation*}
%     2x_{max}-\mu_l-\mu_{l+1} = 0
% \end{equation*}
% Therefore, the maximum point is given by the following equation:
\begin{equation*}
    x_{max} = \frac{\mu_l+\mu_{l+1}}{2}
\end{equation*}
Substituting the value of $x_{max}$ in $D_n$, we obtain:
\begin{equation*}
    \operatorname{sup}_x |F_l(x) - F_{l+1}(x)| = \operatorname{erf}\bigg(\frac{\mu_l - \mu_{l+1}}{2\sqrt{2}\sigma}\bigg) 
\end{equation*}
\end{IEEEproof}}
Based on Lemma \ref{lem:lem3}, we derive the lower bound on $n_T$ given by the following result.

{\begin{lemma}
Given the estimate of the ECDF $F_{\mathcal{S}_e}$ and $F_{\mathcal{S}_t}$ are accurate, for a false-alarm probability of $\mathcal{P}_F$, the minimum number of samples of $\rho_{l+1}$ required to limit the probability of missed detection to $\mathcal{P}_M$ is lower bounded as:
\begin{equation}
    n \geq n_T = \frac{\ln\frac{2}{\mathcal{P}_F}}{2\operatorname{erf}^2\left(\frac{1}{2\sqrt{2}\sigma}\left(\mathcal{P}_M\left(\Delta_{\max} - \Delta_{\min}\right)+\Delta_{\min}\right)\right)}
\end{equation}
\end{lemma}}
\begin{IEEEproof}
The algorithm results in a false-alarm if it detects a change even if the test set and the estimate set are both sampled from the same reward distribution with CDF $F_l(x)$. By the DKWM inequality~\cite{massart1990tight}, for $t_{ref} > 0$, we have:
{
\begin{align}
    \mathbb{P}\left({\mbox{False Alarm}}\right) &= \mathbb{P}\left( \sup_x|F_{\mathcal{S}_t}(x) - F_l(x)| > t_{ref} \big| \mathcal{S}_t \sim \rho_l \right) \nonumber \\
    &\leq 2  \exp\left(-2  n  t_{ref}^2\right)
\end{align}}
%\begin{align}
%    \mathbb{P}\left({\mbox{False Alarm}}\right) &= \mathbb{P}\left(n \cdot sup_x|F_{\mathcal{S}_t} - F_l(x)| > t_{ref} \big| \mathcal{S}_t \sim \rho_l \right) \nonumber \\
%    &\leq 2  \exp(-2t_{ref}^2)
%\end{align}}
%\begin{align}
%    \mathbb{P}\left({\mbox{False Alarm}}\right) &= \mathbb{P}(\sqrt{n}  sup_x|F_{\mathcal{S}_t} - F_l(x)| > t_{ref}|\mathcal{S}_t \sim \rho_l) \nonumber \\
%    &\leq 2  \exp(-2t_{ref}^2)
%\end{align}
{
Equating this to $\mathcal{P}_F$, we get:
\begin{align}
t_{ref} = \sqrt{\frac{1}{2n} \ln \frac{2}{\mathcal{P}_F}}     \label{eq:tref}
\end{align}}
%Equating this to $\mathcal{P}_F$, we get:
%\begin{align}
%t_{ref} = \frac{1}{2}\sqrt{\ln \frac{2}{\mathcal{P}_F}}     \label{eq:tref}
%\end{align}
Recall that a missed detection occurs when the algorithm fails to detect a change, given that the test set and the estimate set are both sampled from two different reward distributions. From the condition of failure of change-point detection:
{
\begin{align}
    &\mathbb{P}\left(\mbox{Missed Detection}\right) \\ &= \mathbb{P}\left( \sup_x|F_{\mathcal{S}_t}(x) - F_l(x)| < t_{ref}|\mathcal{S}_t \sim \rho_{l+1}\right) \nonumber \\
    & = \mathbb{P}\left(\operatorname{erf}\left(\frac{\Delta}{2\sqrt{2}\sigma}\right) < t_{ref}\right) \nonumber \\
    & = \frac{2\sqrt{2}\sigma \operatorname{erf}^{-1}\left(t_{ref}\right) -\Delta_{min}}{\Delta_{\max} - \Delta_{\min}} \nonumber 
\end{align}}
%\begin{align}
%    &\mathbb{P}\left(\mbox{Missed Detection}\right) \\ &= \mathbb{P}\left(\sqrt{n}  sup_x|F_{\mathcal{S}_t} - F_l(x)| < t_{ref}|\mathcal{S}_t \sim \rho_{l+1}\right) \nonumber \\
%    & = \mathbb{P}\left(\sqrt{n}  \operatorname{erf}(\frac{\Delta}{2\sqrt{2}\sigma}) < t_{ref}\right) \nonumber \\
%    & = \frac{\operatorname{erf}^{-1}\left(\frac{t_{ref}}{\sqrt{n}}\right)2\sqrt{2}\sigma -\Delta_{min}}{\Delta_{\max} - \Delta_{\min}} \nonumber 
%\end{align}
The last step follows from the uniform distribution of $\Delta$. Then, the result follows from the condition that $\mathbb{P}\left(\mbox{Missed Detection}\right) \leq \mathcal{P}_M$.
\end{IEEEproof}	
\begin{figure}
    \centering
    \includegraphics[width = 0.8\linewidth]{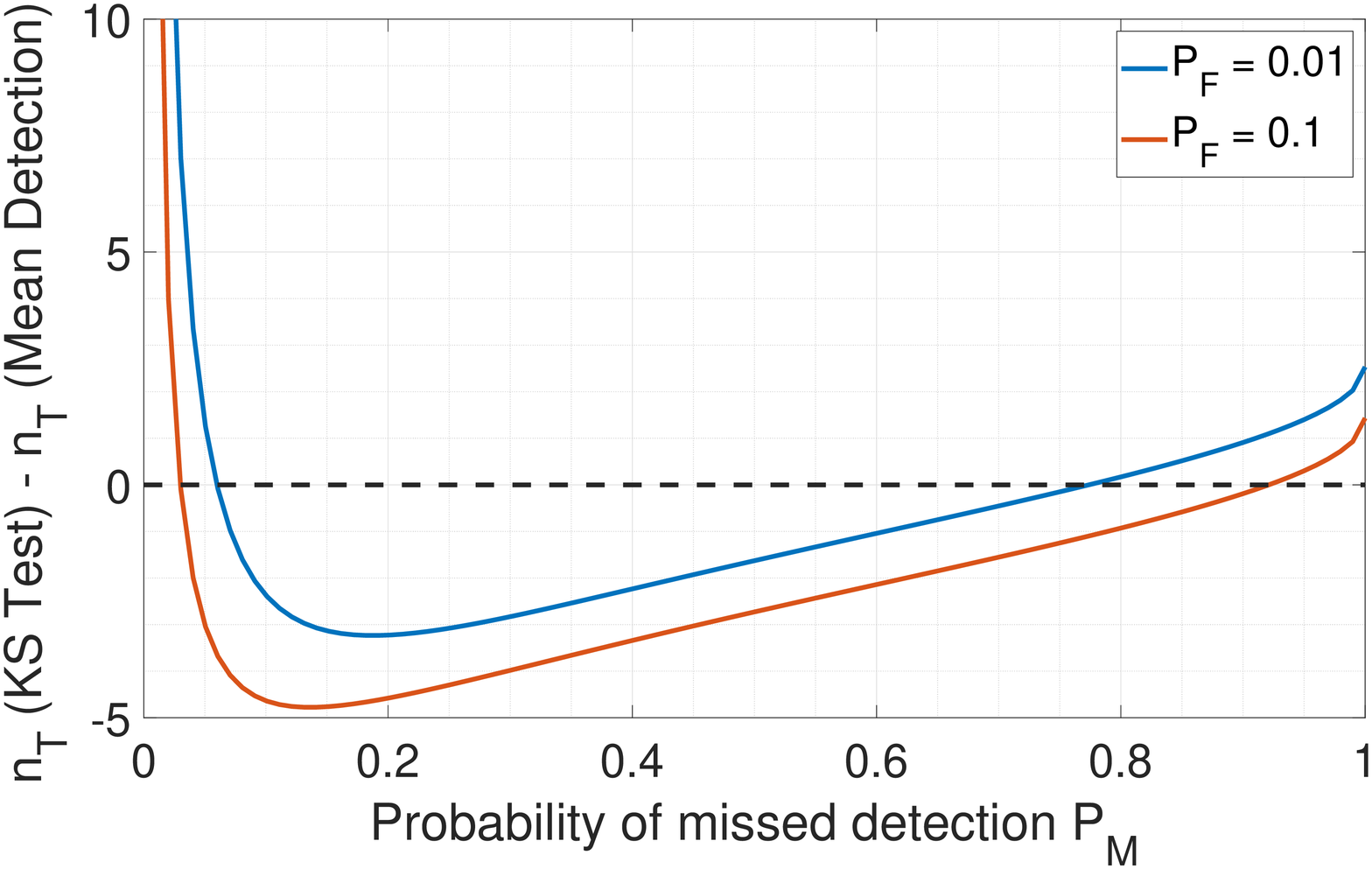}
    \caption{Comparison of the change-detection delay of \texttt{TS-CD} and \texttt{TS-KS}.}
    \label{fig:nT}
    \vspace{-0.8cm}
\end{figure}
{In Fig.~\ref{fig:nT}, we compare the performance of the mean-estimation based change detection method \cite{9194367} with the KS-test based change detection method for the special case when the reward distribution is associated with a change in the mean reward of the arms.} We plot the {difference in the number of samples required to detect a change between the two methods. In particular, we assume the distribution of the reward to be $\sim \mathcal{N}(\mu, \sigma)$ before the change point and $~\mathcal{N}(\mu + \Delta_{min}, \sigma)$ after the change point. Then, the plot in the Figure 1 represents the difference in the number of samples after the change point that the two algorithms \texttt{TS-KS} and \texttt{TS-CD} take to detect the change. Thus, a negative value indicates that \texttt{TS-KS} is able to detect the change point with a fewer samples as compared to \texttt{TS-CD}.}. For very low values of $\mathcal{P}_M$ (e.g., $<0.05$), the mean estimation based method detects a change with fewer samples as compared to the KS-test. {This is because} for such low values of $\mathcal{P}_M$, {the KS-test requires a very high number of samples in order to create an accurate estimate of the CDF}. On the other hand, for a large range of moderately tolerable missed detection probabilities (e.g., 0.1 to 0.7), the number of samples required by the KS-test is fewer than the mean based detection. {It is thus important to note that the proposed KS-test must be employed on top of the bandit framework based on the type of application/service and its requirements characterized by a given missed detection probability. For stringent missed-detection constraints, the mean-estimation based change-detection should be employed since outperforms the KS-test.}

Next, we study the bounds on the change arrival frequency so as to enable the \texttt{TS-KS} algorithm to track the changes in the environment.
\begin{lemma}
To limit the probability of the frequency of change to $p_{change}$, the bound on the value of $\lambda_A$ is:
\begin{align}
    \lambda_A \leq  \frac{1}{n_T + T_N} \ln\left(\frac{1}{1-p_{change}}\right).
\end{align}
\end{lemma}
This follows from \eqref{eq:change}. Finally, in what follows, we derive the bound on the regret of the \texttt{TS-KS} for given values of $\mathcal{P}_M$, $p_{loc}$, and $p_{change}$. For that, we first recall the result of~\cite{agrawal2012analysis} which states that for the stationary regime, the regret the \texttt{TS} algorithm is bounded by the order of $\mathcal{O}(\ln(T))$. Based on this, we derive the following result.
\begin{theorem}
For the two-armed non-stationary bandit problem, with a probability 
\begin{align}
    p_{tot} =  \left(1 - p_{loc}\right)\left(1 - p_{change}\right)\left(1 - \mathcal{P}_M\right),
\end{align}
 the \texttt{TS-KS} algorithm has expected regret bound:
\begin{align}
    \mathbb{E}\left[\mathcal{R}(T)\right]  \leq \mathcal{O}\left(\ln\left(T_N\right) \lambda_A T \left[\frac{\Gamma\left(\frac{T}{T_N + n_T}, \lambda_A T\right)}{\Gamma\left(\frac{T}{T_N + n_T}\right)}\right]\right),
\end{align}
in time $T$, {where $\Gamma(\cdot)$ is the Gamma function, and $\Gamma_\mathrm{i}(\cdot,\cdot)$ is the upper incomplete Gamma function.} Thus, the time expected regret is asymptotically bounded, i.e.,
$
    \lim_{T \to \infty} \frac{1}{T}\mathbb{E}\left[\mathcal{R}(T)\right] = 0.  
$
\end{theorem}
\begin{IEEEproof}
The proof follows similar to that in~\cite{9194367}.
\end{IEEEproof}

\begin{table}
\caption{Reference for Notations used}
\label{tab:Ref}
\centering
\begin{tabular}{ m{2cm} m{6cm} } \toprule
\textbf{Symbol} & \textbf{Definition}\\ \midrule
$T$ & Total number of time-steps/ Time horizon\\ \hline
$\Tilde{r_t} \in \mathbb{R}$ & Reward observed \\ \hline
$x_t \in \{0, 1\}$ & Outcome of the Bernoulli trail \\ \hline
$T_{C_l}$ & Time-step of $l^{th}$ change-point \\ \hline
$F_l(x), F_{l+1}(x)$ & CDF before/after change-point \\ \hline
$F_n(x)$ & Empirical CDF formed from n samples \\ \hline
$D_n$ & Test statistic value obtained from K-S test\\ \hline
$\mathcal{S}_k$ & Reward cache for $k^{th}$ arm \\ \hline
$t_{ref}$ & Reference test statistic value  \\ \hline
$T_N$ & No. of plays of the stationary MAB problem for $N$ plays of the optimal arm\\ \hline
$n_T$ & No. of samples to detect a change \\ \hline
$\mathcal{P}_F, \mathcal{P}_M$ &  Upper limit of the probability of false alarm and the probability missed detection\\
\bottomrule
\end{tabular}
\end{table}

In order to investigate practical applications of our proposed method, in the next sections we perform two case-studies: edge-computing and algorithmic trading.

%\newpage

\section{Case Study 1: Task Offloading in Edge-Computing}
We consider an edge computing framework consisting of multiple edge servers denoted by the set $\mathcal{S}$ and multiple primary users denoted by the set $\mathcal{V}$. Each primary user connects to a server to offload a computing task. Additionally, we consider secondary users which opportunistically associate to a server for offloading their tasks based on the availability of the servers. The primary user-server association changes with time, thereby altering the short-term characteristics of the servers in terms of the amount of task pending to be processed, which we refer to as the \textit{workload buffer}~\cite{dao2020self}. We model the task offloading process of a secondary-user in this network as the \ac{MAB} problem. A secondary user makes a computation request to a server, that must perform the computation within the personal deadline of the secondary user, for the transaction to be counted as a successful service.

An {\it epoch} is defined as the duration for which the number of primary users associated to a server remains constant. Thus, the time axis is divided into sets of epochs, which we denote by $\mathcal{T}_l \equiv \left[ T_{C_l} , T_{C_{l+1}} \right) \cap \mathbb{N} $, where $l \in \mathbb{N}$ and $T_{C_l}$ denotes the $l^{\rm th}$ change-point. For an epoch $\mathcal{T}_l$, the size of the workload buffer at the beginning of this epoch, for server $s_k \in \mathcal{S}$ is $B_{0,l}^{k}$. Furthermore, the servers have a memory limit and accommodate a maximum workload buffer of size $B_{\max}^k$. The number of users associated with server $s_k \in \mathcal{S}$ in epoch $\mathcal{T}_l$ is $m_l^k$, and each primary user offloads task at the same rate of $\eta$ cycles/second. Finally, we assume that the server $s_k$ processes the workload at the rate of $C_k$ cycles/second.

%\subsection{Workload Buffer Evolution}
The size of the workload buffer is modeled as:
\begin{equation}
    B_l^k(t) = B_{0,l}^{k} + (m_l^k \cdot \eta - C_k)\, (t- T_{C_l}); \quad t \in \mathcal{T}_l.
\end{equation}
%\begin{figure}[h!]
%    \centering
%    \includegraphics[width= 0.9\columnwidth]{illustration/two_cases_uniform.eps}
%    \caption{Workload Buffer evolution within an epoch %\textcolor{red}{update notations}}
%    \label{fig:workload}
%\end{figure}
Accordingly, conditioned on $m_l^k$, we model the value of $B_l^k(t), \forall t \in T_{C_i}$, from one of the following distributions:
 \begin{equation*}
    B_{l}^k(t) \sim
    \begin{cases} 
       \mathcal{U}\left( B_0^{l,k} + (m_l^k \cdot \eta -C_k)\, |\mathcal{T}_l|, \quad B_0^{l,k}\right), & m_l^k \eta \leq C_k \\
      \mathcal{U}\left( B_0^{l,k}, \quad B_0^{l,k} + (m_l^k\cdot  \eta -C_k)\, |\mathcal{T}_l|\right), & m_i^k \eta > C_k
   \end{cases}
\end{equation*}
where $|\mathcal{T}_l| \triangleq T_{C_{l+1}} - T_{C_l}$.

\subsection{User-Server association model}
Next, we describe the primary user-server association model. Let us assume that a user $v_j \in \mathcal{V}$ changes its server association at intervals defined by an exponential process with parameter $\lambda_j$ (this model is used in e.g., \cite{rahman2020online}). Then, the interval for which the association of all the primary users remains unchanged is given by the epoch:
\begin{equation}
    |\mathcal{T}_l| \sim  \lambda_C \cdot \exp \left( - \lambda_C x \right); \quad \lambda_C \triangleq \sum_{j=1}^{|\mathcal{V}|} \lambda_j, \quad x\in \mathbb{N}.
    \label{epoch}
\end{equation}
Since the users do not change their server associations within an epoch, we count the number of users associated $m_l^k \geq 0$ with any server $s_k$ by partitioning $\mathcal{V}$ users into $|\mathcal{S}|$ number of disjoint sets, i.e., $\sum_{s_k \in \mathcal{S}} m_l^k = |\mathcal{V}|, \forall l \in \mathbb{N}$.
\begin{equation}
 \mathbf{m}_l \triangleq \Big[m_l^1, m_l^2, \cdots m^{|\mathcal{S}|}_l \Big] \sim \texttt{Partition}(\mathcal{V}, |\mathcal{S}|).
\end{equation}
Where, the \texttt{Partition} mechanism is be modelled as a multinomial distribution:
\begin{equation}
    p_{\mathbf{m}_l}(\mathbf{m}_l = \mathbf{q}_l) = \frac{|\mathcal{V}|!}{\prod_{k=1}^{|\mathcal{S}|} q_l^k! } |\mathcal{S}|^{-|\mathcal{V}|},
\end{equation}
where each primary user is equally likely to connect to a server, i.e., probability of associating with a server $s_k$ is $\frac{1}{|\mathcal{S}|}$, which is independent of the server $s_k$.

\subsection{Task Offload Mechanism and Reward Characterization}
A secondary user wants to offload a task which takes $\delta$ cycles to process. Based on the workload buffer, the time required by server $s_k$ to generate the output is calculated as:
 \begin{equation}
     \zeta_l^k(t) = \frac{B_l^k(t) + \delta}{C_k}
 \end{equation}
For the a task offloading success, the output must be generated within the personal deadline $T_{\max}$ of the secondary user. Thus, the reward for \texttt{TS} is described as:
\begin{equation}
    R_{s_k}(t) = \mathrm{1} \left\{  \zeta_l^k(t) \leq T_{\max} \right\}, \quad \forall t \in \mathcal{T}_l
\end{equation}

% The server which gives the maximum reward within an epoch:
% \begin{align*}
%     s^*[i] &= \arg \max_{s_k \in \mathcal{S}} \mathbb{E}[\rho_k(t)]\\
%     &= \arg \max_{s_k \in \mathcal{S}} \texttt{P}_s(i,k) 
% \end{align*}
For consistency with \texttt{TS-KS}, we map the reward in the continuous domain. Accordingly, we input the value $T_{\max} - \zeta_l^k(t)$ for Algorithm 2. This is intuitive, since negative values indicate that the deadline is exceeded and a positive value refers to the task being completed within the deadline.

\begin{table}[h!]
\caption{Simulation parameters}
\label{tab:Sim}
\centering
\begin{tabular}{ m{2cm} m{3cm} m{2cm} } \toprule
\textbf{Parameter} & \textbf{Value} & \textbf{Unit}\\ \midrule
$B_{\max}^k$        & $\sim \mathcal{U}(0.5, 1)$    & Giga-cycles \\
$C_k$               & $\sim \mathcal{U}(2, 4)$  & GHz\\
$\eta$ & 8 & Mbps\\
$\delta$ & 20 & Mega-cycles\\
$1/\lambda_C$ & $[50, 500]$ & s\\
$T_{\max}$ & 0.05 & s\\
\bottomrule
\end{tabular}
\end{table}

\subsection{Results and Discussion}
\begin{figure}[h!]
    \centering
    \includegraphics[width = 0.8\linewidth]{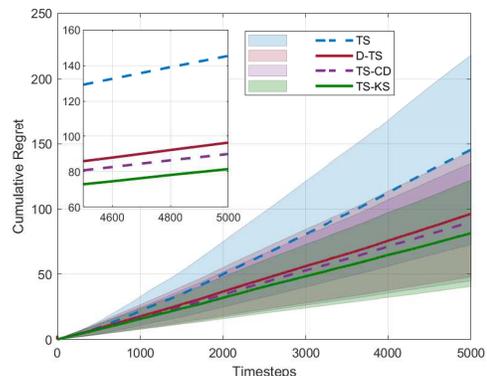}
    \caption{Cumulative Regret. $|\mathcal{S}|=3, |\mathcal{V}|=10^3$.}
    \label{fig:norm_reg_edge}
    \vspace{-0.3cm}
\end{figure}

In Fig.~\ref{fig:norm_reg_edge}, we show the evolution of the cumulative regret with time, for the various bandit algorithms. It is evident that \texttt{TS-KS} beats all other algorithms, as it has the lowest cumulative regret value. The performance of the change-point refresh based algorithms, namely \texttt{TS-CD} and \texttt{TS-KS} is comparable and both have similar variance. In contrast \texttt{TS} not only has the highest regret, but also the highest variance, which makes it the least reliable algorithm for non-stationary environments.

\begin{figure}[h!]
    \centering
    \includegraphics[width = 0.8\linewidth]{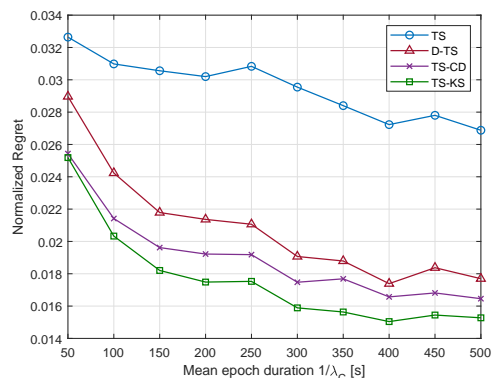}
    \caption{Regret variation against mean epoch duration. $|\mathcal{S}|=3, |\mathcal{V}|=10^3, \frac{1}{\lambda_C}=300~s$.}
    \label{fig:epoch_var}
    \vspace{-0.3cm}
\end{figure}

Next, in Fig.~\ref{fig:epoch_var} we plot the values of mean normalized regret, for different values of mean epoch durations ($\frac{1}{\lambda_C}$) across the bandit algorithms under consideration. The value of mean epoch duration directly translates to the randomness of the environment; If the mean epoch duration is less, the reward distribution of the arms changes quickly and vice versa. 
We see that \texttt{TS-KS} maintains the lowest value of normalized regret across all values of mean epoch duration. As a common trend for all the algorithms, the mean normalized regret decreases as the environment becomes less random, i.e., the mean epoch duration increases.

\section{Case Study 2: Portfolio Optimization}
\textcolor{black}{In the second case-study, we consider the problem of portfolio optimization~\cite{black1992global}. A portfolio is defined as a collection of stocks which are selected on the basis of their associated risks and returns. We employ our \texttt{TS-KS} framework on the New York Stock Exchange dataset~\cite{dataset} and compare it with the contending algorithms. The dataset contains daily prices of 501 stocks spanning from 2010 to the end of 2016. The portfolios considered are $p_{1}$ = \{Eaton Corporation, Whole Foods Market\},  $p_{2}$ = \{Union Pacific Corporation, IDEXX Laboratories\}, $p_{3}$ = \{Union Pacific Corporation\} and $p_{4}$ = \{Church \& Dwights\}. Then, we model each of the portfolios from the set $\mathcal{P} \equiv \{p_1, p_2, p_3, p_4 \}$ as an {\it arm} of the \ac{MAB} setting. For this case-study, we focus only on buying stocks, and accordingly, all the reward returns are functions of unrealized profits in the stocks.
}

\begin{figure}
    \centering
    \includegraphics[width = 0.8\linewidth]{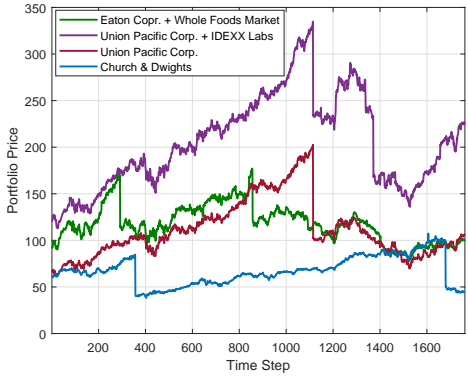}
    \caption{Portfolio price vs Time-step}
    \label{fig:price}
    \vspace{-0.3cm}
\end{figure} 

\textcolor{black}{In Fig.~\ref{fig:price} we show the temporal evolution of the prices of each portfolio. The time-step has an unit of {\it days}. A {\it time-window} is defined as the interval, measured in days, between two consecutive investment actions by the player. The investor periodically chooses a  portfolio $p_k \in \mathcal{P}$ after a gap of $L$ days and invests a certain amount, which is capped at the value $H$. The quantity of stocks in portfolio $p_k$ bought at the $i^{\rm th}$ investment is calculated as $m(i) = \left\lfloor \frac{H}{V^{[k]}(Li)} \right\rfloor$, where $V^{[k]}(t)$ is the combined value of portfolio $p_k$ on the $t^{\rm th}$ day. Furthermore, for the bandit framework, the reward at $i^{\rm th}$ investment is denoted by $\mathcal{R}(i)$ and defined mathematically as:
{
\begin{equation}
    \mathcal{R}(i) \triangleq  \frac{V^{[k]}(Li) \sum_{j \in \mathcal{I}_k} m(j) }{  \sum_{j \in \mathcal{I}_k} V^{[k]}(Lj) m(j)} - 1 ,
\end{equation}}
where $m(j)$ denotes the quantity of stocks in portfolio $p_k$ bought at the $j^{\rm th}$ investment and  $\mathcal{I}_k$ is the set of the investments when portfolio $p_k$ was chosen. The reward is {positive}, when the current valuation of the the stocks of portfolio $p_k$ amassed by the investor is greater than the total investment on it else the reward is {negative}. {For updating the parameters of TS-based algorithms, the reward is translated to binary 1 and 0 for positive and negative values of $\mathcal{R}(i)$ respectively.}}

\begin{figure}
    \centering
    \includegraphics[width = 0.8\linewidth]{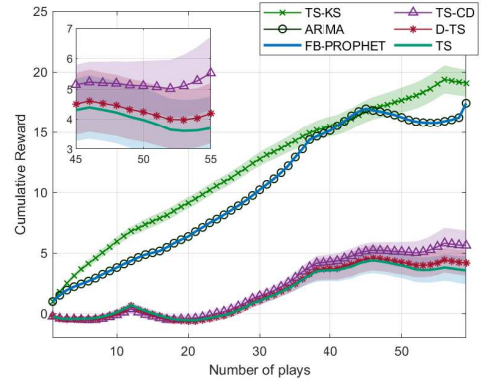}
    \caption{Cumulative reward vs number of plays [time window size = 30 days]}
    \label{fig:cumulative_reward}
    \vspace{-0.3cm}
\end{figure}
\begin{figure}
    \centering
    \includegraphics[width = 0.8\linewidth]{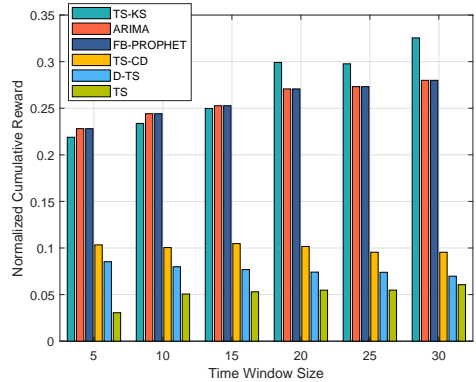}
    \caption{Time normalized cumulative reward vs time window size [in days]}
    \label{fig:time_norm_reward}
    \vspace{-0.6cm}
\end{figure}

{We compare the performance of the proposed \texttt{TS-KS} algorithm with the different variants of Thompson sampling such as \texttt{TS}, \texttt{dTS} and \texttt{TS-CD}. {In addition, we also compare \texttt{TS-KS} with state-of-the art forecasting algorithms such as \texttt{ARIMA}~\cite{al2011modelling} and \texttt{Facebook-PROPHET}~\cite{taylor2018forecasting}.}

{In particular, the ARIMA model considers that the output is linearly dependent on its own previous values and on an imperfectly predictable stochastic term. We select it for comparison since it is typically applied in cases where the data show evidence of non-stationarity in the sense of mean. Recently, in an attempt to develop a model that could capture seasonality in time-series data, Facebook developed the famous \texttt{Facebook-PROPHET} model that is publicly available. It is able to capture daily, weekly and yearly seasonality along with holiday effects, by implementing additive regression models. In essence, \texttt{Facebook-PROPHET} uses a piecewise linear model for trend forecasting. Its fitting procedure is usually very fast (even for thousands of observations) and it does not require any data pre-processing. It deals also with missing data and outliers. Due to its widespread popularity in time-series predictions, we select it as another competitor in our evaluation.}

To facilitate the comparison, first, we consider a fixed window-size of 30 days and plot in Fig.~\ref{fig:cumulative_reward} the temporal evolution of the reward over a period of about 5 years.
Clearly, the TS-KS algorithm outperforms all the contending algorithms. In Fig.~\ref{fig:cumulative_reward}, we see a dip in the baseline algorithms (\texttt{Facebook-PROPHET} and \texttt{ARIMA}) during the phase when the portfolio with the highest valuation crashes (see Fig.~\ref{fig:price}). On the other hand, TS-KS is able to detect the change-points precisely and therefore has uniformly increasing cumulative reward.}
\textcolor{black}{Then, from Fig.~\ref{fig:time_norm_reward}, we observe that the normalized cumulative reward is the maximum for \texttt{TS-KS} across all possible window-sizes varying from {a week to a month}. This establishes the efficacy of the proposed \texttt{TS-KS} algorithm vis-a-vis \texttt{TS-CD} and \texttt{dTS}.}

{Currently, we are investigating risk-constrained bandit algorithms in non-stationary environments, which we will report in a future work.}

\section{Conclusions and Future Work}
We have proposed a KS-test based change detection mechanism for non-stationary environments in the MAB framework. \textcolor{black}{Based on the KS-test, we have proposed an actively adaptive TS algorithm called \texttt{TS-KS} for the MAB problem. For the two-armed bandit case, the \texttt{TS-KS} algorithm has a sub-linear regret. It is noteworthy that the proposed framework is able to detect a change when the state-of-the-art mean-estimation based change detection mechanisms fail. Leveraging two case-studies, we have demonstrated that the \texttt{TS-KS} algorithm outperforms not only the classical \texttt{TS} and passively adaptive \texttt{D-TS} algorithms, but also it performs better than the actively adaptive \texttt{TS-CD} algorithm.}
{Moreover, the performance of \texttt{TS-KS} is at par with the state-of-the-art forecasting algorithms such as \texttt{Facebook-PROPHET} and \texttt{ARIMA} as demonstrated in the portfolio optimization case study.}

\bibliographystyle{IEEEtran}
\bibliography{main.bib}
		
\end{document}